\documentclass[10pt,twocolumn,letterpaper]{article}
\usepackage[T1]{fontenc}
\usepackage{lmodern}
\usepackage[margin=0.8in,columnsep=0.28in]{geometry}
\usepackage{graphicx}
\usepackage{amsmath,amssymb}
\usepackage{booktabs}
\usepackage{xcolor}
\usepackage[numbers,sort&compress]{natbib}
\usepackage{enumitem}
\setlist[itemize]{noitemsep,leftmargin=*,topsep=0pt}
\setlist[enumerate]{noitemsep,leftmargin=*,topsep=0pt}
\newcommand{\etal}{\emph{et al.}}

\definecolor{linkblue}{rgb}{0.21,0.49,0.74}
\usepackage[breaklinks,colorlinks,allcolors=linkblue]{hyperref}
\usepackage[capitalize]{cleveref}
\crefname{section}{Sec.}{Secs.}
\Crefname{section}{Section}{Sections}
\crefname{table}{Tab.}{Tabs.}
\Crefname{table}{Table}{Tables}

\title{BoviShift: A Six-Year Multi-Attribute Audit of Visual Cattle Phenotyping Across Animals and Time}

\author{%
Mobasshir Bhuiya Shagor\\
Graaho Technologies LLC\\
Dhaka, Bangladesh\\
{\tt\small mobasshir.bhuiya@graaho.com}
\and
Nilambar Halder Tonmoy\\
Indiana University Bloomington\\
Bloomington, IN, USA\\
{\tt\small nhaldert@iu.edu}
\and
Md Saef Ullah Miah\\
American International University Bangladesh\\
Dhaka, Bangladesh\\
{\tt\small saef@aiub.edu}
}

\hypersetup{
  pdftitle={BoviShift: A Six-Year Multi-Attribute Audit of Visual Cattle Phenotyping Across Animals and Time},
  pdfauthor={Mobasshir Bhuiya Shagor; Nilambar Halder Tonmoy; Md Saef Ullah Miah}
}
\date{}

\begin{document}
\maketitle

\begin{abstract}
Computer vision offers a scalable, non-contact means of estimating livestock
traits, but correlated images and video frames make reliable evaluation
difficult. Image-random splitting can place views of the same animal in both
training and test sets, while deployment further requires transfer to animals
captured in future years. We introduce BoviShift, a corpus reconstructed from
six consecutive annual cohorts of an operational cattle marketplace collected
from 2021 to 2026. It contains 2{,}657 cattle, 10{,}610 still images, and 2{,}566
walk-around videos across nine recorded breed categories. We evaluate three
splitting strategies through four configurations: image-random and pooled
cattle-disjoint splits over the same development cohorts, followed by separate
future-2025 and locked future-2026 evaluations. Using a fixed ResNet-50
multi-task baseline, we assess weight, age, height, breed, and coat color.
Differences between image-random and cattle-disjoint splitting are
task-dependent rather than a universal leakage penalty. Future-cohort behavior
is also attribute- and cohort-specific: breed declines relative to pooled
evaluation; mean weight and age performance fails to beat constant baselines
in 2025; height remains above baseline; and color transfers well to 2025 but
degrades in 2026. Multi-task training improves all five point estimates in a
single-seed ablation, still-view aggregation helps selectively, and zero-shot
video-frame aggregation exposes a still-to-video domain gap. To our knowledge,
BoviShift is the longest-spanning cattle-vision resource combining consecutive
annual cohorts with physical and visual attributes. Upon acceptance, we will release the corpus with
its 79{,}390 sampled video frames, alongside reproducible
metadata reconstruction, data-quality audits, leakage-checked split manifests,
and evaluation code for realistic, repeatable visual livestock phenotyping.
\end{abstract}

\section{Introduction}
\label{sec:intro}
Estimating livestock traits from images could reduce reliance on manual
measurement and specialized equipment. Weighing cattle conventionally requires
a crush, scale, and trained handler, whereas photographs can be acquired with a
phone. This distinction matters where weighing infrastructure is unavailable
and sale decisions depend on visual judgment. Recent methods report promising
errors for image-based weight estimation~\cite{lawe2025,pointnet2023,depthcalf2025}.

Their interpretation, however, depends on whether observations of the same
animal can occur in both training and test sets.

Livestock datasets rarely contain only one frame per animal. A marketplace
listing may include four still photographs and a walk-around video captured in
the same session and setting. An image-level partition can therefore place
views of the same animal on both sides of the split.

The model is then evaluated on new observations, but often of animals already
represented during training. This identity overlap is a common source of
over-optimistic evaluation in applied machine learning~\cite{kapoor2023}.

It may allow a model to exploit animal- or capture-specific cues rather than
generalizable phenotype evidence.

A deployed phenotyping system must also process future animals after population
and acquisition conditions have changed. Benchmarks assembled during one
collection campaign cannot directly test this setting because their partitions
share a common acquisition period. Although temporal distribution shift is well
studied elsewhere~\cite{wildtime2022,wilds2021}, it has received little
systematic study across multi-year livestock-vision cohorts.

We study both problems using six consecutive annual cohorts of a commercial
cattle marketplace collected from 2021 to 2026. BoviShift contains 2{,}657
cattle, 10{,}610 still images, 2{,}566 walk-around videos, and 79{,}390 sampled
frames across nine recorded breed categories, with no cattle identifier occurring in more than one cohort.

The design is therefore repeated cross-sectional rather than longitudinal:
annual differences may combine changes in population, acquisition, and
annotation rather than within-animal growth.

We evaluate three splitting strategies through four configurations. P1 randomly
partitions images from the 2021--2024 development cohorts, whereas P2 partitions
the same cohorts by cattle identity. P3 trains on 2021--2023, validates on 2024,
and tests on 2025. The locked P4 run trains through 2024, validates on 2025, and
tests once on 2026. Because partition units, test identities, and cohort
composition change across configurations, their gaps are operational
differences rather than causal effects of identity leakage or elapsed time.

Using a fixed ResNet-50 model family, we evaluate weight, age, height, breed,
and coat color.

Future-cohort performance is attribute- and cohort-dependent: breed declines;
the mean weight and age results fail to beat constant baselines in 2025;
height remains above baseline; and color transfers well to 2025 but declines in
2026. We also compare single- and multi-task training, still-view aggregation,
and zero-shot aggregation of sparse video frames.

Our contributions are:

\begin{enumerate}
\itemsep0.15em
\item \textbf{BoviShift}, a six-year corpus of 2{,}657 cattle with multi-view
stills, walk-around videos, and five physical and visual attributes, released in
full on acceptance.

\item \textbf{A four-configuration evaluation framework} comparing image-random,
pooled cattle-disjoint, and two future-cohort settings using leakage-checked
manifests, cattle-level metrics, and simple baselines.

\item \textbf{A reproducible integrity audit} that repairs 9{,}295 media
references and identifies size-dependent missing weight and price labels in
40\% of the 2024 cohort.
\end{enumerate}

\section{Background and Related Work}
\label{sec:background}

\subsection{Visual estimation of livestock traits}

Body-weight estimation from imagery has a long lineage in precision livestock
farming, and the best-performing systems are consistently those with the most
controlled capture. Point-cloud approaches using depth sensors report mean
absolute percentage errors near 3\% on herds of around one hundred
animals~\cite{pointnet2023}, and depth-image pipelines for dairy calves reach
comparable accuracy~\cite{depthcalf2025}. More
recently, lightweight networks have brought the task to commodity hardware: one
such system reports accuracy above 97\% for weight estimation from a mobile
phone~\cite{lawe2025}.

These are real results, but the evaluation base is narrow. The mobile-phone
system above is validated on 108 animals of a single breed; the point-cloud work
on roughly 100. Neither the effect of breed heterogeneity nor the effect of
elapsed time between training and deployment can be assessed at that scale,
because the variation simply is not present in the data. Whether a 97\% figure
survives a change of breed, season or year is, at present, an open question
rather than a settled one. Part of our purpose here is to make it answerable.

\subsection{Cattle image datasets}

Public cattle datasets have grown substantially, but they have grown along the
identity axis rather than the phenotype axis. BECA~\cite{beca2025} is the largest
recent release, with 16{,}889 images of 5{,}661 beef cattle assembled for
long-term individual recognition; its long-term subset follows 103 animals for up
to five months, a span its authors describe as unusually long for the field.
MOO~\cite{moo2026} contributes multi-view observations for viewpoint analysis in
re-identification. Both are strong resources for recognizing \emph{which} animal
is in frame. Neither carries body-weight, height or price labels, and neither
spans multiple years.

On the phenotype side, CID~\cite{cid2022} released 17{,}899 images of 513 cattle
with weight, height and breed annotations for joint regression and
classification, drawn from a single collection campaign. A recent survey of
public computer vision datasets for precision livestock farming~\cite{plfsurvey}
catalogues the wider landscape and reaches a similar conclusion: datasets are
either large and identity-focused, or trait-annotated and small.

BoviShift sits in the gap. It is trait-annotated, comparable in image count to
the identity datasets, released in full on acceptance, and --- unlike any of the
above ---
organized into six consecutive annual cohorts drawn from one continuously
operating marketplace.

\subsection{Reconstruction-based approaches}

A parallel line of work estimates traits by first recovering geometry.
Agreement-driven fusion of multi-view segmentations with a 3D foundation model
reports $R^2 = 0.69$ and MAPE of 2.22\% for live cattle
weight~\cite{multiview3d2026}, and monocular mesh recovery has been applied to
body measurement in goats~\cite{goat2026}. The recent SAM 3D
Animal~\cite{sam3danimal2026} makes promptable animal reconstruction in the wild
practical, building on a SMAL-family parametric model.

\citet{multiview3d2026} report that the bottleneck for multi-view 3D reconstruction is data: producing 3D-suitable capture at scale on working farms is difficult. BoviShift contributes 2,566 walk-around videos collected during routine marketplace operations using a simple handheld acquisition process. Although these are not calibrated 3D captures, their scale and viewpoint variation provide a useful resource for developing and evaluating reconstruction methods under less controlled conditions. The SAM 3D Animal model of \citet{sam3danimal2026} builds on the SMAL family, which recovers shape only up to an unknown scale and was fitted predominantly to taurine and general quadruped morphology. Whether it faithfully represents the thoracic hump and dewlap of Bos indicus cattle is not established, and a metric weight estimate cannot be obtained without an explicit calibration step. We therefore treat reconstruction as a direction the corpus enables rather than a baseline we adopt uncritically.
\subsection{Evaluation protocols, leakage and temporal shift}

The methodological framing of this paper draws on work outside livestock vision.
WILDS~\cite{wilds2021} assembled distribution shifts that arise in deployment
rather than in construction, and Wild-Time~\cite{wildtime2022} extended this to
shift over time, reporting substantial degradation between in-distribution and
temporally out-of-distribution evaluation across domains including healthcare and
remote sensing. Neither includes an agricultural or livestock benchmark.

On the leakage side, Kapoor and Narayanan~\cite{kapoor2023} document how
train--test contamination produces irreproducible results across a range of
applied fields, and subject-level splitting is standard practice in medical
imaging for precisely this reason. Livestock vision has not generally adopted the
convention, in part because many datasets contain one image per animal and the
question does not arise. Once multi-view stills and video frames are introduced,
it does.

What is missing in both literatures is a setting where the two effects can be measured on the same data, with the same models, at the same time.
That is what
the cohort structure of BoviShift permits.

\section{The BoviShift Corpus}
\label{sec:dataset}

\subsection{Source and collection}

BoviShift extends CID~\cite{cid2022} with five further annual cohorts. The 2021
cohort corresponds to the collection released as CID, recovered here as 514
animals after re-indexing; images, videos, and cattle
properties for the remaining cohorts were sourced directly from the Bengal Meat
marketplace.
Each listing represents one animal and typically includes
seller-recorded attributes, four still photographs captured against a prepared
backdrop, and a handheld walk-around video. The six annual cohorts span
2021--2026, and the marketplace operator granted written permission for
redistribution (\Cref{sec:availability}).

The corpus represents a narrow production setting: 2,646 of 2,657 animals are intact male beef-finishing bulls, with six heifers and five steers. Accordingly, results should not be extrapolated to dairy cattle, breeding stock, or other regions without additional validation. Recorded health fields are above 99.8\% positive and are not used as targets. No cattle identifier recurs across cohorts. BoviShift is therefore treated as repeated cross-sectional rather than longitudinal: annual differences combine changes in sampled population, acquisition, and annotation, not
within-animal growth.
\begin{table*}[t]
\centering
\small
\setlength{\tabcolsep}{5pt}
\begin{tabular}{@{}lrrrrrrr@{}}
\toprule
 & 2021 & 2022 & 2023 & 2024 & 2025 & 2026 & Total \\
\midrule
\multicolumn{8}{@{}l}{\emph{Media}} \\
Cattle & 514 & 629 & 399 & 385 & 344 & 386 & 2{,}657 \\
Still images & 2{,}056 & 2{,}516 & 1{,}596 & 1{,}537 & 1{,}370 & 1{,}535 & 10{,}610 \\
Walk-around videos & 512 & 621 & 378 & 384 & 288 & 383 & 2{,}566 \\
Sampled video frames & 15{,}843 & 19{,}155 & 11{,}687 & 11{,}904 & 8{,}928 & 11{,}873 & 79{,}390 \\
Animals without video & 2 & 8 & 21 & 1 & 56 & 3 & 91 \\
\midrule
\multicolumn{8}{@{}l}{\emph{Labels}} \\
Usable weight labels & 510 & 629 & 399 & 231 & 344 & 385 & 2{,}498 \\
Weight $\mu$ (kg) & 256.4 & 271.7 & 294.5 & 254.4$^{\dagger}$ & 270.2 & 288.0 & 272.9 \\
Weight $\sigma$ (kg) & 74.3 & 81.2 & 107.4 & 83.4 & 92.1 & 94.5 & 89.4 \\
Height $\mu$ (in) & 46.7$^{\ddagger}$ & 42.4 & 43.0 & 42.6 & 42.7 & 43.3 & 43.5 \\
Median price (BDT k) & 101 & 139 & 160 & 152 & 156 & 161 & --- \\
\midrule
\multicolumn{8}{@{}l}{\emph{Composition} (\%)} \\
LOCAL & 68.9 & 81.2 & 78.2 & 93.0 & 87.5 & 75.8 & 80.1 \\
SAHIWAL & 11.5 & 10.5 & 14.0 & 2.1 & 1.5 & 13.5 & 9.3 \\
SINDHI & 7.4 & 1.3 & 0.3 & 0.0 & 0.0 & 1.0 & 1.9 \\
RED coat color & 59.9 & 59.0 & 53.6 & 56.1 & 50.9 & 49.2 & 55.5 \\
\bottomrule
\end{tabular}
\caption{The corpus by cohort. Media totals cover 92{,}566 objects and 37.3\,GB.
Label statistics exclude records flagged as missing (\cref{sec:missing}).
$^{\dagger}$2024 weight is biased downward by non-random missingness.
$^{\ddagger}$2021 height reflects a measurement-convention change and is not
comparable with later cohorts.}
\label{tab:corpus}
\end{table*}

\subsection{Media inventory}
\label{sec:media-inventory}

\Cref{tab:corpus} summarizes the corpus. Coverage is close to uniform: 2{,}651 of
2{,}657 animals have exactly four still images, and the remaining six have one.
Video coverage is 96.6\% overall but uneven by year. In 2025, 288 of 344 cattle
have an archived video, leaving 56 without one. After still-image eligibility
filtering, the P3 test set contains 342 cattle, of which 288 have video; this is
the 54-cattle shortfall reported in the video experiment. Video results are
therefore restricted to the covered subset, with its sample size stated.

\paragraph{Frame extraction.} The released video frames are sampled at a fixed count of 31 frames per video with random timing, spanning the full duration of each video. Frames are stored as JPG and match the source video resolution. We remove frames flagged as corrupt or truncated during decoding, yielding 79,390 usable frames across 2,566 videos (mean 30.9 frames per video). This release-time extraction is independent of the eight-frame evaluation protocol in \Cref{sec:multi-ablations}, which draws a separate, sparser set of frames at fixed relative timestamps for that specific experiment.

\begin{table}[t]
\centering
\footnotesize
\setlength{\tabcolsep}{3.5pt}
\begin{tabular}{@{}llll@{}}
\toprule
Attribute & Target & Output & Metric \\
\midrule
Weight & continuous & kg & MAE \\
Age & ordered numeric & months & MAE \\
Height & continuous & inches & MAE \\
Breed & categorical & 6 classes & macro-F1 \\
Color & categorical & 2 classes & macro-F1 \\
\bottomrule
\end{tabular}
\caption{Benchmark targets and primary metrics. Regression mean and
classification majority baselines are fit separately on each training split.}
\label{tab:attributes}
\end{table}

\subsection{Attributes}

We predict five attributes (\cref{tab:attributes}). Weight and height are
continuous, with ranges of 140--826~kg and 30--65~inches after validity checks.
Age is recorded in five ordered levels from 24 to 48 months and is treated as
regression. It closely follows seller-recorded incisor count, so it should be
interpreted as an estimated dental-age category rather than a verified birth
record. Missing or invalid continuous targets are masked per task.

Breed is highly imbalanced: LOCAL accounts for 80.1\% of labeled cattle, while
PABNA, MIR KADIM, DHANNI, and BRAHMA contain six or fewer each. Categories with
fewer than 20 training cattle are pooled into \textsc{Other}, yielding six
evaluation classes, and performance is reported with macro-F1. Color is mapped
to \textsc{Red} versus \textsc{Non-Red}; the three unsupported
\textsc{Cross-Red} records are masked. Its relative visual simplicity makes
color a useful control for acquisition and cohort effects.

\subsection{Metadata reconstruction and integrity audit}

The corpus as originally archived was not usable as a benchmark, and repairing it
was a substantial part of this work. We report the process because the resulting
manifests are part of what we release, and because the failure modes are likely
to recur in any dataset assembled incrementally over several years.

Four defect classes were resolved. Space normalization repaired 2{,}263 video
and 7{,}032 image references, and 64{,}131 of 92{,}566 objects were brought to
one path convention. The missing 2026 index was reconstructed for 386 cattle.
Six misnamed videos were reconciled using source identifiers and hashes, one
byte-identical video was deduplicated, and one conflicting metadata row was
removed. The audit also found 30 still-image rows in 13 exact-hash groups; these
files are documented, and any hash associated with multiple cattle IDs is
excluded before benchmark splitting.

The final media index contains 92{,}566 paths joined to annual metadata by
normalized cattle identifier, with no unmatched image records. Checksums and
byte sizes permit verification after re-hosting; path occurrence should not be
confused with unique file content because the duplicate-hash audit is retained.

\begin{table}[t]
\centering
\footnotesize
\setlength{\tabcolsep}{4pt}
\begin{tabular}{@{}lrrr@{}}
\toprule
Stratum & Labeled & Missing & Difference \\
\midrule
size $=$ MINIMUM & 88.2\% & 72.1\% & $-16.1$ pt \\
size $=$ MEDIUM & \phantom{0}8.7\% & 23.4\% & $+14.7$ pt \\
size $=$ EXTRA\_LARGE & \phantom{0}0.4\% & \phantom{0}3.2\% & $+2.8$ pt \\
mean height (in) & 42.1 & 43.1 & $+1.0$ in \\
\midrule
$n$ & 231 & 154 & 385 total \\
\bottomrule
\end{tabular}
\caption{Composition of the 2024 cohort split by label availability.
\emph{Labeled}: nonzero weight and price. \emph{Missing}: weight and price recorded as zero. \emph{Difference}: Missing minus Labeled, in percentage
points \textit{(pt)} for the size shares and inches \textit{(in)} for mean height. The MINIMUM share is 16.1\,\textit{pt} lower among the unlabeled animals, while
MEDIUM and EXTRA\_LARGE are 14.7 and 2.8\,\textit{pt} higher, and mean height is
1.0\,\textit{in} greater. Missingness is therefore size-dependent, with the unlabeled
animals systematically larger. Because larger animals weigh more, the 2024 mean
weight computed over the labeled subset alone (\cref{tab:corpus}) is biased
downward.}
\label{tab:missing}
\end{table}

\subsection{Missingness and the 2024 cohort}
\label{sec:missing}

Among the 2{,}657 deduplicated cattle, 159 non-positive weight and price entries
are treated as missing rather than imputed; 154 occur in 2024 and affect 40.0\%
of that cohort. \Cref{tab:missing} shows size-dependent missingness: unlabeled
cattle contain fewer \textsc{Minimum} and more \textsc{Medium} or
\textsc{Extra-Large} records, and are one inch taller on average. The observed
2024 mean weight of 254.4\,kg is therefore biased downward.

Invalid labels contribute neither loss nor metrics. Accordingly, we do not select 2024 as a future evaluation cohort, and its observed weight is not interpreted as evidence of
temporal shift. It serves as P3 validation data and enters P4 training with
missing targets masked. P3 contains 1{,}540 eligible training cattle from
2021--2023, of which 1{,}538 have valid weight labels. The manifests also mask
the implausible 4.6-inch height record and document three cattle with incomplete
media and obsolete source URLs.

\subsection{Cohort composition drift}

The shift the corpus is built to measure is present in the labels before any
model is trained, and the lower panels of \cref{tab:corpus} make it concrete.

Breed composition moves substantially. The LOCAL share ranges from 68.9\% to
93.0\%; SAHIWAL falls from 14.0\% in 2023 to 1.5\% in 2025 before returning to
13.5\% in 2026; SINDHI declines from 7.4\% in 2021 to complete absence in 2024
and 2025. Four of the nine breed categories appear in only one or two cohorts. A
future-year protocol on this corpus therefore encounters class disappearance
and reappearance, not merely reweighted frequencies. The \textsc{Red} share
declines overall from 59.9\% to 49.2\%, with a temporary increase in 2024.

Prices rise, as expected in a nominal currency over a six-year window: the median
listing moves from 101{,}000 to 161{,}000, a nominal increase of roughly 59\%.
This matters for any economically labeled target, and it is the reason we do not
treat price as a phenotype in this study. Price is nonetheless released with the
corpus, since it supports questions this paper does not attempt.

Two discontinuities indicate acquisition change. Mean recorded height falls
from 46.7 inches in 2021 to 42.4 inches in 2022 and remains between 42.4 and
43.3 thereafter, suggesting a change in measurement convention; 2021 height is
therefore flagged as not directly comparable. The feed schema also changes
after 2022 from an ingredient list to one categorical type. Both changes are
documented in the datasheet and exemplify the operational drift the temporal
protocols are designed to expose.

\subsection{Benchmark organization}

We instantiate three splitting strategies through four configurations
(\cref{tab:protocols}). P1 and P2 use the 2021--2024 development cohorts with
image-random and cattle-disjoint partitions, respectively. P3 trains on
2021--2023, validates on 2024, and tests on 2025; locked P4 trains through 2024,
validates on 2025, and tests once on 2026. P2--P4 are cattle-disjoint. Explicit
manifests record cattle IDs and content hashes, and the verification script
rejects cross-partition identity or exact-file overlap. Protocol contrasts are
descriptive because partition units and test populations change together.

\subsection{Ethics, licensing and availability}
\label{sec:availability}

\noindent\textbf{Ethics.}
The corpus derives from commercial listings published by the marketplace operator – Bengal Meat. Seller and buyer identifiers are not retained, and the benchmark contains no associated personal records. No animal was handled, restrained, or otherwise affected for this research; the marketplace produced the imagery in the ordinary course of trade.

\noindent\textbf{Permission and license.}
Redistribution rights for the imagery, video, derived frames, and attribute
metadata were granted in writing by the marketplace operator for research use.
The specific license terms, including any media-specific restrictions stated
separately for stills, videos, and derived frames, will accompany the release.
The operator is acknowledged in the camera-ready version in accordance with the
terms of that agreement.

\noindent\textbf{What is released.}
The release package comprises 10{,}610 still images, 2{,}566 walk-around videos,
and 79{,}390 sampled frames (extraction protocol detailed in \Cref{sec:media-inventory}; 92{,}566 objects, 37.3\, GB); attributes for 2{,}657 cattle; the checksummed media index; audit and exclusion records; manifests for all four configurations; leakage verification; and baseline training and
evaluation code. Upon acceptance, it will be deposited in a persistent public
archive with a DOI and no registration requirement. A datasheet following Gebru
\etal~\cite{gebru2021} documents collection, composition, preprocessing,
recommended uses, and limitations.

\noindent\textbf{Reproducibility.}
The reported dataset summaries and evaluations can be regenerated from the
release without contacting the authors. Split manifests contain explicit cattle
identifiers rather than only a random seed, allowing future implementations to
reuse the same partitions and verify identity overlap directly.

\noindent\textbf{Intended use and limitations.}
Automated appraisal can shift bargaining power between buyers and sellers, and
subgroup error should be audited before deployment. This study does not validate
a commercial valuation system. The corpus covers one marketplace, one country,
and one narrow production category; its intended use is research on visual
phenotyping, evaluation methodology, and distribution shift. Deployment in other
populations requires local validation.

\section{Evaluation Protocol}
\label{sec:evaluation}

We examine how image-random splitting affects measured performance, how
cattle-disjoint models transfer to future cohorts, and whether multi-task and
multi-observation inference improve that transfer. We hold the model family
fixed, but interpret protocol differences descriptively because partition unit
and test-cohort composition change together.

\subsection{Leakage and temporal protocols}
\label{sec:protocols}

\Cref{tab:protocols} defines four configurations. P1 randomly assigns images
from 2021--2024; consequently, 899 of 913 test cattle also occur in training.
P2 uses the same cohorts and approximately 70/15/15 proportions but partitions
by cattle ID. P3 trains on 2021--2023, validates on 2024, and tests on 2025;
the locked P4 run trains on 2021--2024, validates on 2025, and tests once on
2026. P2--P4 have no cattle overlap between partitions. Before splitting, we
remove 30 images whose content hash occurs under multiple cattle IDs; no
remaining file hash crosses a partition. This leaves 10,580 eligible images,
with 2,647 of 2,657 cattle represented. All manifests and
preprocessing rules are fixed before training.

\begin{table*}[t]
  \centering
  \small
  \setlength{\tabcolsep}{5pt}
  \caption{Evaluation configurations. Each split reports images/cattle. P1 is
  intentionally leakage-prone; P2--P4 are cattle-disjoint.}
  \label{tab:protocols}
  \begin{tabular}{llccc}
    \toprule
    Config. & Split rule & Train & Validation & Test \\
    \midrule
    P1 & image-random, 2021--2024
       & 5,382/1,908 & 1,153/919 & 1,153/913 \\
    P2 & cattle-disjoint, 2021--2024
       & 5,381/1,347 & 1,156/289 & 1,151/288 \\
    P3 & future 2025
       & 6,152/1,540 (2021--2023) & 1,536/384 (2024) & 1,368/342 (2025) \\
    P4 & locked future 2026
       & 7,688/1,924 (2021--2024) & 1,368/342 (2025) & 1,524/381 (2026) \\
    \bottomrule
  \end{tabular}
\end{table*}

Breeds represented by fewer than 20 training cattle are mapped to
\textsc{Other}, yielding six common evaluation classes from the nine recorded
categories. Color is evaluated as \textsc{Red} versus \textsc{Non-Red}; weight,
age, and height retain their recorded units. Unsupported categorical and invalid
continuous labels are masked. Label maps and normalization statistics are fit
on training data only.

\subsection{Model and optimization}
\label{sec:model-training}

Our reference model uses an ImageNet-pretrained ResNet-50 encoder~\cite{resnet}
followed by dropout ($p=0.2$), one linear classification head for each of breed
and color, and a joint linear head for the three continuous attributes. Regression
targets are standardized using the training-set mean and standard deviation.
The multi-task objective is the equally weighted sum of (i) the mean
inverse-frequency-weighted cross-entropy over the two classification tasks,
with weights computed from training image rows, and (ii) Smooth-$L_1$ loss over
all valid standardized regression targets. Thus, the classification and
regression components each have coefficient one; no test-set metric is used to
tune their relative weighting.

More precisely, for classification tasks $\mathcal{C}$, regression tasks
$\mathcal{R}$, standardized target $z_{ir}$, and validity mask $m_{ir}$, we
optimize
\begin{equation}
\begin{aligned}
\mathcal{L}={}&\frac{1}{|\mathcal{C}|}
\sum_{c\in\mathcal{C}}\operatorname{CE}_{w_c}(y_c,\hat{y}_c) \\
&+\frac{\sum_{i,r}m_{ir}\,
\operatorname{SmoothL1}(z_{ir},\hat{z}_{ir})}
{\sum_{i,r}m_{ir}}.
\end{aligned}
\label{eq:multitask-loss}
\end{equation}
For residual $e=\hat{z}-z$, Smooth-$L_1$ with $\beta=1$ uses
$e^2/2$ when $|e|<1$ and $|e|-1/2$ otherwise. It is quadratic for small
errors but less sensitive than squared error to large outliers. Because targets
are standardized, the transition occurs at one training-set standard deviation.
The mask ensures that an unavailable continuous target contributes neither loss
nor a regression metric; unsupported categorical targets are ignored by
cross-entropy.

The shorter image side is resized to 256 pixels for a $224\times224$ input.
Training uses a random resized crop (scale 0.75--1.0, aspect ratio 0.75--1.33),
horizontal flipping with probability 0.5, and color jitter with brightness,
contrast, saturation, and hue magnitudes of 0.15, 0.15, 0.10, and 0.02; evaluation uses a center crop. We optimize all parameters for 30 epochs with
AdamW, batch size 64, initial learning rate $3\times10^{-4}$, weight decay
$10^{-4}$, gradient-norm clipping at 1.0, cosine learning-rate decay, and
bfloat16 mixed precision. The
checkpoint with minimum total validation loss is selected without inspecting
the test set. P1--P3 are run with seeds 42, 123, and 2026. P4 uses the frozen
configuration and seed 42 as a single pre-designated held-out run. The three
P1--P3 seeds vary model initialization and training order, not the fixed split
manifest.

\subsection{Metrics and uncertainty}
\label{sec:metrics}

Breed and color are evaluated with macro-F1; weight, age, and height use MAE in
kilograms, months, and inches. Cattle-level predictions average class
probabilities or regression outputs across available stills. We report mean and
sample standard deviation across seeds where available. Conditional 95\%
percentile intervals use 2,000 cattle-bootstrap resamples, with paired resampling
for ablations. They capture test-cohort sampling uncertainty for a fitted
checkpoint, not retraining, split, or cohort-selection variation. Training-cattle
majority and mean predictors provide classification and regression baselines.

P1 contains 1.26 test views per animal on average, whereas P2 keeps approximately
four; the two configurations also contain different test identities. We
therefore use image-level metrics only for the descriptive P1--P2 split-policy
comparison and cattle-level metrics for P2--P4. The P1--P2 contrast is an
operational difference between splitting policies, not a causal estimate of
identity leakage.

\subsection{Multi-observation and task ablations}
\label{sec:multi-ablations}

\paragraph{Single- versus multi-task learning.}
On P3 with seed 42, we train five single-task models, one per attribute. Each
uses the same encoder, split, augmentation, optimizer, epoch budget, and
checkpoint rule as the multi-task model. This tests joint supervision with a
shared representation for the seed-42 training configuration.

\paragraph{Still-view aggregation.}
We compare all-view cattle predictions with one randomly selected still. The
single-view point estimate averages 2,000 random selections; each paired
bootstrap replicate resamples cattle and draws a new view, avoiding dependence
on one favorable selection.

\paragraph{Sparse video frames.}
We test whether the frozen still-image model transfers directly to the
walk-around videos. In contrast to the 79,390 release frames (\Cref{sec:media-inventory}), we sample eight candidate frames at evenly spaced relative timestamps from 10\% to 90\% of each video. The one-frame set uses 44.3\%, while the eight-frame set uses all candidates. Predictions are aggregated
by averaging class probabilities and regression outputs. Still--video comparisons use exactly the cattle with an available, successfully decoded video: 288/342 in P3 and 381/381 in P4. This is
zero-shot still-to-video transfer; no video frame is used for training or model
selection.

\section{Results}
\label{sec:results}

\subsection{Image-random versus cattle-disjoint splitting}
\label{sec:leakage-results}

Table~\ref{tab:leakage-results} compares P1 and P2 at the image level, avoiding the
different number of test views in their cattle-level aggregates. Relative to
P1, P2 has 0.060 lower breed macro-F1 and 1.575~kg and 0.131-month higher weight
and age MAE. The direction is not universal: P2 has 0.011 higher color F1 and
0.195-inch lower height MAE. Weight also varies substantially across seeds. We
therefore do not report a single ``leakage penalty.'' These are descriptive
differences between fixed configurations with matched years and similar data
budgets, but different test identities.

\begin{table}[t]
  \centering
  \scriptsize
  \caption{Image-level P1--P2 comparison, reported as mean $\pm$ sample
  standard deviation over three optimization seeds. Higher is better for F1;
  lower is better for MAE. MAE units are kg, months, and inches.}
  \label{tab:leakage-results}
  \resizebox{\columnwidth}{!}{%
  \begin{tabular}{lccccc}
    \toprule
    Config. & Br. F1 $\uparrow$ & Col. F1 $\uparrow$ & Wt. MAE $\downarrow$
      & Age MAE $\downarrow$ & Ht. MAE $\downarrow$ \\
    \midrule
    P1 image-random
      & $0.500\!\pm\!0.029$ & $0.885\!\pm\!0.020$
      & $35.256\!\pm\!3.173$ & $2.291\!\pm\!0.040$ & $1.948\!\pm\!0.393$ \\
    P2 cattle-disjoint
      & $0.440\!\pm\!0.026$ & $0.896\!\pm\!0.011$
      & $36.831\!\pm\!12.075$ & $2.422\!\pm\!0.206$ & $1.753\!\pm\!0.314$ \\
    \bottomrule
  \end{tabular}
  }
\end{table}

\subsection{Cattle-level future-cohort evaluation}
\label{sec:main-results}

\Cref{tab:main-results} uses the animal as the evaluation unit for the
cattle-disjoint and future-cohort configurations. Relative to P2, P3
is associated with 0.067 lower breed F1 and 30.591~kg, 0.423-month, and
0.725-inch higher MAE for weight, age, and height. Color remains similar
($0.910$ versus $0.920$). This is an operational configuration gap, not a causal
estimate of calendar time: P2 randomly partitions 2021--2024, whereas P3
trains on 2021--2023, validates on 2024, and tests on a different 2025
population.

\begin{table}[t]
  \centering
  \scriptsize
  \caption{Cattle-level performance. Model rows for P2--P3 are mean $\pm$
  sample standard deviation over three optimization seeds; P4 is one locked
  run. Base denotes the training-cattle majority/mean predictor. Weight, age,
  and height MAE use kg, months, and inches.}
  \label{tab:main-results}
  \resizebox{\columnwidth}{!}{%
  \begin{tabular}{lccccc}
    \toprule
    Config. & Br. F1 $\uparrow$ & Col. F1 $\uparrow$ & Wt. MAE $\downarrow$
      & Age MAE $\downarrow$ & Ht. MAE $\downarrow$ \\
    \midrule
    P2 model
      & $0.469\!\pm\!0.037$ & $0.910\!\pm\!0.007$
      & $33.836\!\pm\!13.255$ & $2.401\!\pm\!0.219$ & $1.603\!\pm\!0.368$ \\
    P2 base & 0.145 & 0.372 & 53.222 & 2.750 & 2.603 \\
    P3 model
      & $0.402\!\pm\!0.067$ & $0.920\!\pm\!0.008$
      & $64.427\!\pm\!16.399$ & $2.824\!\pm\!0.279$ & $2.328\!\pm\!0.471$ \\
    P3 base & 0.234 & 0.340 & 61.056 & 2.790 & 3.375 \\
    P4 model & 0.312 & 0.657 & 38.915 & 2.969 & 1.815 \\
    P4 base & 0.144 & 0.329 & 64.536 & 3.020 & 3.437 \\
    \bottomrule
  \end{tabular}
  }
\end{table}

The large weight standard deviations in P2 and P3 reveal optimization
sensitivity. The P3 seed-42 checkpoint reaches 45.936~kg MAE, whereas the
three-seed mean (64.427~kg) is worse than the 61.056~kg training-mean baseline;
the mean age result similarly fails to improve on its baseline. In contrast,
P3 breed, color, and height and all five P4 point estimates improve on their
corresponding baselines. However, the P3 height baseline is inflated by the
2021 measurement-convention change: including that cohort raises the training
mean, so the margin over baseline narrows when 2021 is excluded. We retain the
cohort and report the baseline explicitly rather than adjusting it. These
outcomes make the seed distribution and simple baselines essential parts of the
benchmark rather than optional diagnostics.

P4 is a second held-out future cohort rather than a direct repeat of P3: it
adds 2024 to training, uses 2025 for validation, and tests a different 2026
population. Its breed and color F1 are 0.312 and 0.657; conditional 95\%
cattle-bootstrap intervals are [0.252, 0.375] and [0.605, 0.706]. Weight, age,
and height MAE are 38.915~kg, 2.969 months, and 1.815 inches, with conditional
95\% intervals [35.560, 42.415], [2.737, 3.228], and [1.639, 1.990]. The P4
regression rebound alongside the color decline illustrates cohort-specific
behavior, but the single P4 training seed precludes a claim about its
optimization variance.

\subsection{Does multi-task learning help?}
\label{sec:multitask-results}

Table~\ref{tab:single-multi} compares seed-42 models on exactly the same P3
test cattle. Multi-task learning improves every point estimate for this
training seed. Conditional on the fitted seed-42 checkpoints, the paired 95\%
cattle-bootstrap interval excludes zero for weight, age, and height. Breed and
color also favor the multi-task model, but their intervals include zero. Since
only one optimization seed is available for this ablation, these intervals do
not establish that the multi-task advantage persists across retraining.

\begin{table}[t]
  \centering
  \scriptsize
  \caption{P3 single-task versus multi-task learning (seed 42). The final
  column is single-task minus multi-task under paired cattle bootstrap. Thus,
  positive MAE differences and negative F1 differences favor multi-task
  learning.}
  \label{tab:single-multi}
  \resizebox{\columnwidth}{!}{%
  \begin{tabular}{llccc}
    \toprule
    Attribute & Metric & Multi-task & Single-task & Difference [95\% CI] \\
    \midrule
    Breed  & macro-F1     & 0.421  & 0.362  & $-0.059$ [$-0.103$, $+0.074$] \\
    Color  & macro-F1     & 0.915  & 0.904  & $-0.010$ [$-0.046$, $+0.027$] \\
    Weight & MAE (kg)     & 45.936 & 63.603 & $+17.667$ [$+15.325$, $+19.828$] \\
    Age    & MAE (months) & 2.516  & 2.691  & $+0.175$ [$+0.090$, $+0.257$] \\
    Height & MAE (inches) & 1.799  & 1.928  & $+0.129$ [$+0.025$, $+0.233$] \\
    \bottomrule
  \end{tabular}
  }
\end{table}

\subsection{Aggregation across still views}
\label{sec:view-results}

\Cref{tab:view-gains} reports the gain from using all available still
views instead of one randomly selected view. The point estimates show positive
weight, height, and color gains in P2, P3, and P4. The age gains are small,
and breed improves in P2 and P3 but reverses on P4. We therefore treat
multi-view aggregation as task- and cohort-dependent rather than universally
beneficial.

For the locked P4 checkpoint, paired cattle-bootstrap intervals exclude zero
only for the weight gain (+1.539~kg [0.072, 3.048]) and height gain
(+0.115~inches [0.040, 0.186]). The P4 breed, color, and age intervals include zero. These intervals reflect test-cattle resampling for one trained model and do not measure retraining uncertainty.

\begin{table}[t]
  \centering
  \scriptsize
  \caption{Gain from all-view cattle aggregation over a randomly selected
  single still. Positive values favor aggregation. P2 and P3 show mean
  $\pm$ standard deviation over three seeds; P4 is one locked run. MAE units
  are kg, months, and inches.}
  \label{tab:view-gains}
  \resizebox{\columnwidth}{!}{%
  \begin{tabular}{lccccc}
    \toprule
    Config. & Br. F1 & Col. F1 & Wt. MAE & Age MAE & Ht. MAE \\
    \midrule
    P2 & $+0.028\!\pm\!0.018$ & $+0.014\!\pm\!0.014$
       & $+2.994\!\pm\!1.977$ & $+0.023\!\pm\!0.022$ & $+0.150\!\pm\!0.055$ \\
    P3 & $+0.039\!\pm\!0.035$ & $+0.031\!\pm\!0.008$
       & $+1.782\!\pm\!1.117$ & $+0.064\!\pm\!0.041$ & $+0.119\!\pm\!0.069$ \\
    P4 & $-0.041$ & $+0.008$ & $+1.539$ & $+0.009$ & $+0.115$ \\
    \bottomrule
  \end{tabular}
  }
\end{table}

\subsection{Zero-shot transfer to video frames}
\label{sec:video-results}

Table~\ref{tab:video-results} compares uniformly sampled video frames with
curated still views on the matched cattle subset. Eight-frame video aggregation
is worse than all-still aggregation for every P3 metric. The paired intervals
exclude zero for all five differences conditional on the seed-42 checkpoint.
In P4, video is worse for breed, weight, and height; color is worse with eight
frames but better with one; the small age improvement is inconclusive. Eight
frames underperform the predefined one-frame sample on all five P4 point estimates, although the paired age
difference includes zero.

\begin{table}[t]
  \centering
  \scriptsize
  \caption{Zero-shot sparse-frame evaluation using frozen still-image models.
  Still results use the identical video-available cattle subset. MAE units
  are kg, months, and inches.}
  \label{tab:video-results}
  \resizebox{\columnwidth}{!}{%
  \begin{tabular}{llccccc}
    \toprule
    Config. & Input & Br. F1 & Col. F1 & Wt. MAE & Age MAE & Ht. MAE \\
    \midrule
    P3 & one video frame   & 0.307 & 0.775 & 52.272 & 2.869 & 3.003 \\
    P3 & eight video frames& 0.312 & 0.836 & 60.031 & 2.835 & 3.231 \\
    P3 & all still views   & 0.432 & 0.912 & 43.668 & 2.462 & 1.790 \\
    \midrule
    P4 & one video frame   & 0.250 & 0.680 & 48.158 & 2.867 & 2.354 \\
    P4 & eight video frames& 0.174 & 0.596 & 50.933 & 2.903 & 2.507 \\
    P4 & all still views   & 0.312 & 0.657 & 38.915 & 2.969 & 1.815 \\
    \bottomrule
  \end{tabular}
  }
\end{table}

This result does not show that video is intrinsically less informative. Because
the model is trained only on curated stills, the gap may reflect motion blur,
occlusion, unfavorable poses, compression, or other acquisition differences in
uniformly sampled frames. Video-specific training and quality-aware frame
selection remain open benchmark directions.

\section{Discussion}
\label{sec:discussion}

The experiments yield three main observations. First, P1 and P2 produce
task-dependent differences rather than a universal leakage penalty. Because
P1 nevertheless places the same cattle in training and test, it remains an
unsuitable estimate of deployment performance. Second, the future-2025
configuration is markedly harder for breed and the physical attributes than
pooled cattle-disjoint evaluation, while color behaves differently in 2025
and 2026. Generalization is therefore both attribute- and cohort-specific.
Third, the large across-seed variation in weight estimation shows that a
single training run is insufficient evidence of robustness. Additional
observations are also selective: curated still aggregation benefits some
physical attributes, whereas naive video-frame averaging exposes a
still-to-video domain gap.

These results motivate a practical reporting standard: partition by cattle
identity, include a held-out future cohort, report multiple optimization
seeds, aggregate and evaluate at cattle level, and obtain uncertainty by
resampling cattle rather than correlated images. Image-random results may be
reported as a diagnostic, but not as evidence of deployment generalization.

BoviShift also has important limitations. It is reconstructed from one
operational marketplace, so annual differences combine changes in population,
capture practice, operations, and annotation rather than isolating calendar
time. P1--P2 and P2--P3 consequently compare fixed experimental
configurations, not causal effects, and each uses one split realization.
Breed labels are imbalanced, color is binary, and marketplace metadata may
contain measurement or transcription error. Our baseline uses one encoder
family; P4 and the single-task ablation use one training seed, and checkpoint
bootstrap intervals do not capture retraining variation. Finally, video
evaluation is zero-shot, and 54 P3 test cattle lack videos. Future work should
test additional model families, split realizations, and video-specific methods
under the same fixed identity-disjoint temporal evaluation strategy.

{\small
\bibliographystyle{ieeenat_fullname}
\bibliography{main}
}

\end{document}